\documentclass{article}

\usepackage{arxiv}

\usepackage[utf8]{inputenc} 
\usepackage[T1]{fontenc}    
\usepackage{hyperref}       
\usepackage{url}            
\usepackage{booktabs}       
\usepackage{amsfonts}       
\usepackage{nicefrac}       
\usepackage{microtype}      
\usepackage{lipsum}
\usepackage{float}
\usepackage{caption}
\usepackage{graphicx}
\usepackage{multirow}

\title{False Positive Reduction by Actively Mining Negative Samples for Pulmonary Nodule Detection in Chest Radiographs}

\author{
Sejin Park\thanks{Corresponding author: gnoses@gmail.com} \\
VUNO Inc.\\
Seoul, South Korea \\
\And
Woochan Hwang \\
School of Medicine\\
Imperial College London\\
UK \\
\And
Kyu Hwan Jung \\
VUNO Inc.\\
Seoul, South Korea\\
\AND
\\
\And
Joon Beom Seo \\
Asan Medical Center\\
University of Ulsan \\
Seoul, South Korea\\
\And
Namkug Kim \\
Asan Medical Center \\
University of Ulsan \\
Seoul, South Korea\\
}

\begin{document}
\maketitle

\begin{abstract}

Generating large quantities of quality labeled data in medical imaging is very time consuming and expensive. The performance of supervised algorithms for various tasks on imaging has improved drastically over the years, however the availability of data to train these algorithms have become one of the main bottlenecks for implementation. To address this, we propose a semi-supervised learning method where pseudo-negative labels from unlabeled data are used to further refine the performance of a pulmonary nodule detection network in chest radiographs. After training with the proposed network, the false positive rate was reduced to 0.1266 from 0.4864 while maintaining sensitivity at 0.89.

\end{abstract}

\keywords{Convolutional neural network \and Lung nodule detection \and False positive reduction \and Negative sample mining \and Semantic segmentation}

\section{Introduction}

For many medical images, labeled data (with lesion labels) can be in short supply. However, unlabeled data (with unknown lesion and class labels) tends to be more readily available. This has been a major limiting factor in applying machine learning to medical imaging. To tackle this problem, we propose a novel semi supervised learning method for lung nodule detection in chest radiographs.     

While CAD (Computer aided detection) systems can achieve high sensitivity, their relatively low specificity has limited its implementation in the clinical setting. Most CAD systems therefore, utilize various false positive reduction methods to improve system specificity. Our approach here is to use semi-supervised learning, which allows us to use the pseudo-negative labeled data hidden amongst the unlabeled data.

The goal of this study is to develop a method that allows us to use the pseudo-negative labeled data via transfer learning to improve performance of the detection task. We also showed that using clinically verified negative data improves specificity. On the other hand, we found that the inclusion of whole unlabeled data, which has both positive and negative data, degrades the sensitivity score.

\section{Methods}

We used chest radiographs (PA view) collected from 2013 to 2015 at Asan Medical Center. There were 1970 labeled radiographs, where suspicious lung nodules were marked by clinicians on the images as regions of interest. We also had 3000 unlabeled chest radiographs, which were used during the second phase of our proposed method. In addition, we used 3000 clinically verified normal radiographs to validate our hypothesis that using additional negative data will improve the false positive rate. The 1970 radiographs used for this study were anonymized and reviewed by the internal review board at Asan Medical Centre before being provided to us.

During the pre-processing stage, we used per image histogram equalization to mitigate the variance of intensity amongst the radiographs. In our model, data augmentation (rotation, random crop, resize, intensity and contrast noise) had no significant effect in terms of training and validation accuracy. Therefore, we did not apply any augmentations to our data as they increased the training time with no significant benefit.

Our proposed method is a two phase learning model. In the first phase, we used the pixel labeled data from 1970 labeled radiographs to train a U-net [1] like encoder decoder network for semantic segmentation. To handle small sized nodules, a class balanced loss function was used. The U-net like lateral/skip connection from lower layers to higher layers performs well on small datasets of medical images. Then we added inception modules in several layers of the encoder part which made our network 30\% faster without degradation of accuracy. We trained our model for 20 hours using the ADAM [2] optimizer in Tensorflow.

For the first phase, we confirmed our best FROC score from the 1970 labeled data via 5 folds cross validation. The 1970 cases were divided into 5 subsets of similar size. For each fold, 4 subsets were used for training and 1 subset was used for validation and testing.

\begin{figure}[h!]
\centering
\includegraphics[width=100mm]{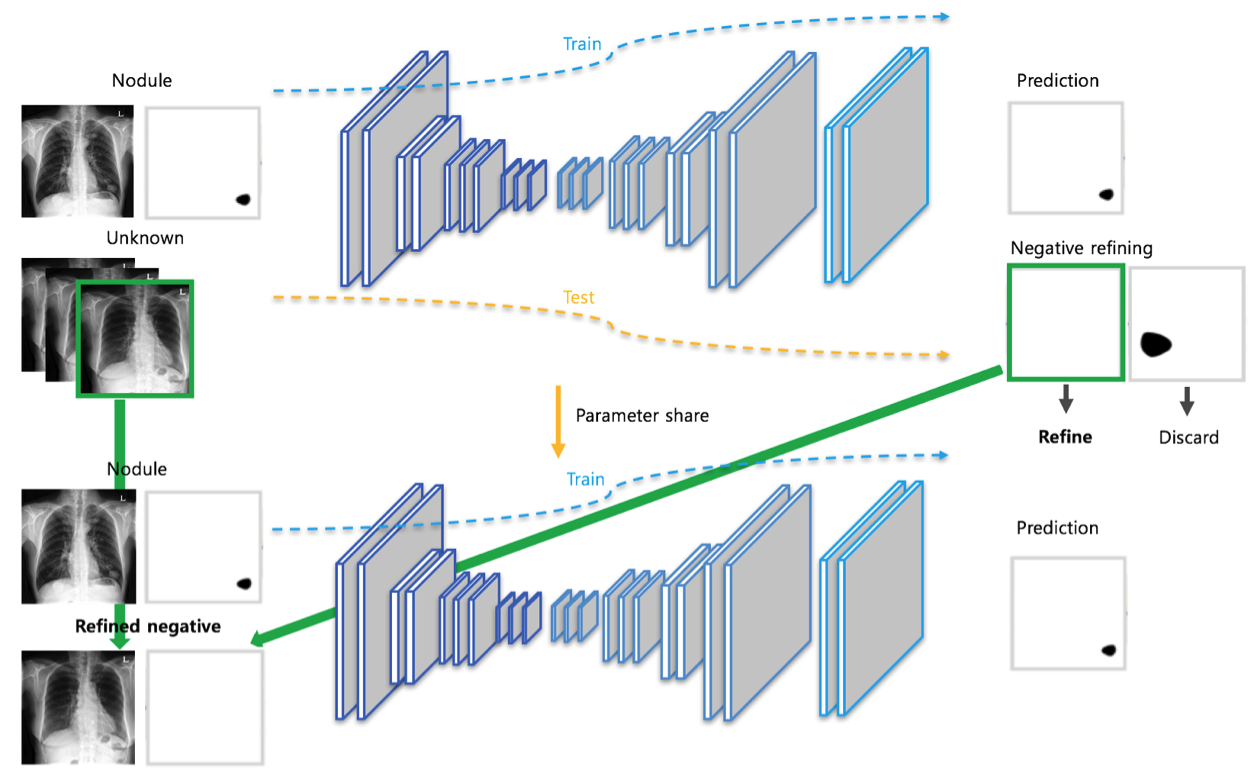}
\caption{Training procedure and network architecture of proposed method}
\label{fig:architecture}
\end{figure}

During the second phase, we used the trained network from the first phase to discover pseudo-negative samples from prediction results of the 3000 unlabeled data. If no nodules were detected from the prediction, the samples were considered pseudo-negative samples. We discarded any other unlabeled data. Both labeled and refined pseudo negative samples were then used jointly to train our model. The network in the second phase had transferred weights from the whole layers of the first phase network. The network architecture and the test set of the first and second phase is identical.

\section{Results}
The proposed ‘Negative sample refining method’ attained a performance of 0.1266 false positives per scan. Our results show that using refined pseudo-negative data can decrease the number of false positives per scan by 73\% while maintaining sensitivity.

As seen in Table \ref{tab:performance}, FROC of the first phase was 0.8924 and 0.4864 for sensitivity and false positives per scan, respectively. After the second phase, the false positive rate is improved to 0.1266, while maintaining baseline sensitivity at 0.889, which is superior to previous studies.

\begin{table}[h]
\centering
\caption{Comparison of FROC performance in first and second phase}
\label{tab:performance}
\centering
\begin{tabular}{l|llllll}
\multicolumn{1}{c|}{\multirow{2}{*}{Fold}} & \multicolumn{2}{c|}{Phase 1}                                     & \multicolumn{2}{c|}{Phase 2}                                     & \multicolumn{2}{c}{Difference}                                      \\
\multicolumn{1}{c|}{}                      & \multicolumn{1}{c|}{Sensitivity} & \multicolumn{1}{c|}{FP / Img} & \multicolumn{1}{c|}{Sensitivity} & \multicolumn{1}{c|}{FP / Img} & \multicolumn{1}{c|}{(Sensitivity)} & \multicolumn{1}{c}{(FP / Img)} \\ \hline
1                                          & 0.894                            & 0.273                         & 0.884                            & 0.113                         & -0.01                              & -0.16                          \\ \hline
2                                          & 0.894                            & 0.508                         & 0.892                            & 0.165                         & -0.002                             & -0.343                         \\ \hline
3                                          & 0.907                            & 0.428                         & 0.895                            & 0.066                         & -0.012                             & -0.362                         \\ \hline
4                                          & 0.902                            & 0.756                         & 0.910                            & 0.154                         & 0.008                              & -0.602                         \\ \hline
5                                          & 0.865                            & 0.467                         & 0.864                            & 0.135                         & -0.001                             & -0.332                         \\ \hline
\multicolumn{1}{|l|}{Avg}                  & \multicolumn{1}{l|}{0.8924}      & \multicolumn{1}{l|}{0.4864}   & \multicolumn{1}{l|}{0.889}       & \multicolumn{1}{l|}{0.1266}   & \multicolumn{1}{l|}{-0.003}        & \multicolumn{1}{l|}{-0.36}     \\ \hline
\end{tabular}
\end{table}

\begin{figure}[h!]
\centering
\includegraphics[width=100mm]{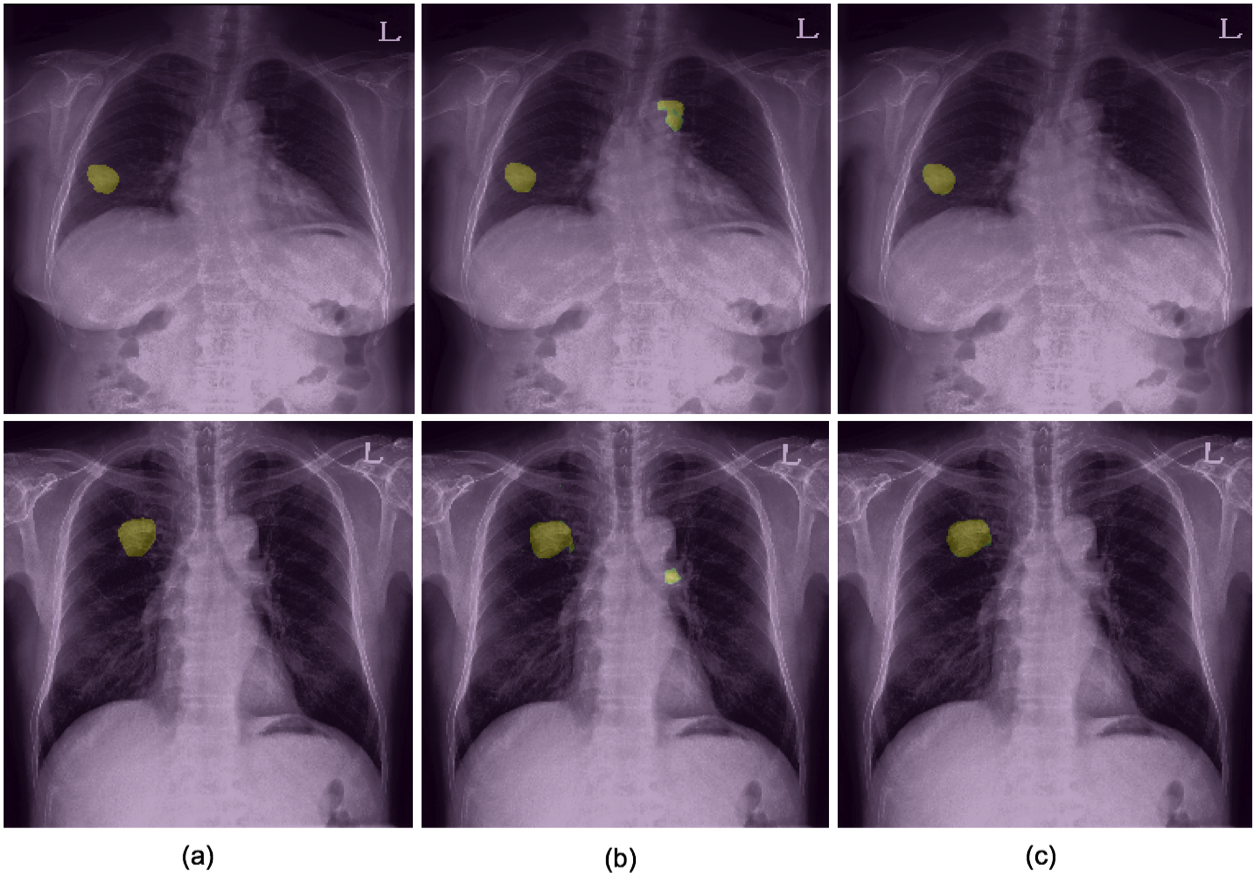}
\caption{Illustration of result by proposed method
(a) gold standard (b) first phase (c) second phase
}
\label{fig:results}
\end{figure}

Figure \ref{fig:results} shows the lesions detected by our model after each phase. In the figure, we can observe that the false positive lesion detected after the first phase (b) disappears after the second phase (c).

In order to prove that discovering true negative samples from unlabeled data is an effective approach to false positive reduction, we experimented by using clinically verified negative data. As seen in Table \ref{tab:performance2}, the sensitivity was maintained at 0.89, while the false positive rate decreased to 0.0827. These results support the hypothesis which led to the development of our semi supervised learning method.

\begin{table}[h]
\centering
\caption{Comparison of FROC performance in 3 data sources. Approved : clinically approved true negative data. Pseudo negative : refined pseudo negative data. Unlabeled : unlabeled data}
\label{tab:performance2}
\centering
\begin{tabular}{|l|l|l|}
\hline
               & Sensitivity & FP / Scan \\ \hline
Approved       & 0.890       & 0.0827    \\ \hline
Pseudonegative & 0.889       & 0.1266    \\ \hline
Unlabeled      & 0.853       & 0.15      \\ \hline
\end{tabular}
\end{table}

Furthermore, we validated our prediction that sensitivity will decrease when second phase learning is performed using the complete set of unlabeled data without refining. In Table \ref{tab:performance2}, the sensitivity of 0.8924 after the first phase decreases to 0.853 after the second phase training with unrefined unlabeled data.

\section{Discussion}

Through this study, we showed that significant improvement in the false positive rate can be achieved by our proposed method. We confirmed that clinically verified true negatives are effective in reducing the false positive rate, and the fact that the resulting false positive rate is similar to what we achieved with refined pseudo-negative data shows the potential of this method.
Because we only looked for the lack of nodules during the selection process however, the refined pseudo negative samples can contain other disease features. Therefore, these samples can not be regarded as true negatives or normal data. Nonetheless, the given task is detecting nodules not all abnormal lesions, and because the data used for phase 1 training were not labeled for non-nodular lesions to start with, the pseudo negative data found by our method is not meaningless. Also, since the experimental results show improvement of the FROC defined for a given problem, the proposed method can be claimed to be effective.

\section{Conclusion}
We proposed a novel semi supervised learning method for false positive reduction of the nodule detection task in chest radiographs. Our method shows significant improvements to the conventional method, and thus provides a practical solution to bringing CAD systems one step closer to clinical implementation.  

\section{References}

[1] Ronneberger, Olaf, Philipp Fischer, and Thomas Brox. "U-net: Convolutional networks for biomedical image segmentation." International Conference on Medical image computing and computer-assisted intervention. Springer, Cham, 2015.

[2] Kingma, Diederik P., and Jimmy Ba. "Adam: A method for stochastic optimization." arXiv preprint arXiv:1412.6980 (2014).

\end{document}